\documentclass[letterpaper, 10 pt, conference]{ieeeconf}

\newcommand{\horizon}{T}
\newcommand{\state}{x}

\newcommand{\control}{u}

\newcommand{\numplayers}{N}
\newcommand{\feedback}{\gamma}
\newcommand{\feedbackset}{\Gamma}
\newcommand{\runningcost}{g}
\newcommand{\cost}{J}

\newcommand{\advrunningcosti}{g_{\textnormal{adv}, i}}
\newcommand{\cooprunningcosti}{g_{\textnormal{coop}, i}}

\newcommand{\advhorizon}{T_{\textnormal{adv}}}

\newcommand{\thresh}{d}
\newcommand{\lane}{\ell}

\newcommand{\xdim}{n}
\newcommand{\udim}{m}

\newcommand{\ra}{\rightarrow}
\newcommand{\R}{\mathbb{R}}

\newtheorem{definition}{Definition}

\newcommand{\figref}[1]{Fig.~\ref{#1}}
\newcommand{\secref}[1]{Sec.~\ref{#1}}

\usepackage{graphicx}
\usepackage{adjustbox}
\usepackage{wrapfig}
\usepackage[labelfont=bf,figurename=Fig.,font=small]{caption}
\usepackage{subcaption}
\usepackage{lipsum}
\usepackage{amssymb}
\usepackage{amsmath}
\usepackage{siunitx}
\usepackage{mathtools}
\usepackage{xcolor}
\usepackage[ruled,vlined,linesnumbered]{algorithm2e}
\usepackage[noend]{algpseudocode}
\usepackage{ifthen}
\usepackage{dsfont}
\usepackage[backend=biber,style=numeric-comp,sorting=none,maxbibnames=99]{biblatex}
\usepackage{balance}
\usepackage{dirtytalk}
\usepackage{afterpage}

\usepackage{times}

\bibliography{books, papers}

\IEEEoverridecommandlockouts
\title{\LARGE \bf Encoding Defensive Driving as a Dynamic Nash Game}

\author{
Chih-Yuan Chiu,$^*$ David Fridovich-Keil,$^*$ and Claire J. Tomlin
\thanks{
$^*$The first two authors contributed equally to this work.}%
\thanks{
C-Y Chiu and C Tomlin are with the Dept. of Electrical Engineering \& Computer Sciences, UC Berkeley. D Fridovich-Keil is with the Dept. of Aeronautics \& Astronautics, Stanford University. Correspondence to \tt \small{david.fridovichkeil@stanford.edu}.}%
\thanks{
This research is supported by an NSF CAREER award, the Air Force Office of Scientific Research (AFOSR), NSF's CPS FORCES and VeHICaL projects, the UC-Philippine-California Advanced Research Institute, the ONR MURI Embedded Humans, a DARPA Assured Autonomy grant, and the SRC CONIX Center.}
}

\begin{document}

\maketitle
\thispagestyle{empty}
\pagestyle{empty}

\begin{abstract}

Robots deployed in real-world environments should operate safely in a robust manner. In scenarios where an \say{ego} agent navigates in an environment with multiple other \say{non-ego} agents, two modes of safety are commonly proposed---adversarial robustness and probabilistic constraint satisfaction. However, while the former is generally computationally intractable and leads to overconservative solutions, the latter typically relies on strong distributional assumptions and ignores strategic coupling between agents.

To avoid these drawbacks, we present a novel formulation of robustness within the framework of general-sum dynamic game theory, modeled on defensive driving. More precisely, we prepend an \textit{adversarial phase} to the ego agent's cost function. That is, we prepend a time interval during which other agents are assumed to be temporarily distracted, in order to render the ego agent’s equilibrium trajectory robust against other agents' potentially dangerous behavior during this time. We demonstrate the effectiveness of our new formulation in encoding safety via multiple traffic scenarios. 

\end{abstract}

\section{Introduction}
\label{sec:intro}

Decision-making modules in autonomous systems must meet safety and robustness criteria before they are deployed in real-world settings with uncertain or unknown environments.
Scenarios such as autonomous driving, in which an \say{ego} agent must interact with other, possibly non-cooperative \say{non-ego} agents, are of particular interest.
These scenarios are naturally modeled using dynamic game theory, which describes the evolution of each agent’s state according to their minimization of a cost function. Each player's cost function depends on the control strategies of that player, as well as the shared state of all the players.

To ensure safe and efficient operation of the ego agent in such multi-agent settings, existing methods formulate safety in the following two ways. Adversarial robustness methods, such as Hamilton-Jacobi-Isaacs (HJI) equation-based reachability theory, aim to generate trajectories that would ensure the safety of the ego agent despite worst-case behavior of all other agents \cite{bansal2017hamilton, mitchell2005time, margellos2011hamilton}. Another commonly proposed methodology involves probabilistic constraint satisfaction \cite{pilipovsky2020ChanceConstrainedOptimal, OnoBlackmore2010ChanceConstrainedFiniteHorizonOptimalControl}. Here, algorithms attempt to bound the probability that the ego agent's trajectory becomes unsafe. 
Unfortunately, each of these approaches carries significant drawbacks. HJI methods only apply to zero-sum game settings, and exact solution methods suffer from the so-called \say{curse of dimensionality,} with computational cost increasing exponentially in the dimension of the state \cite{bansal2017hamilton}. Meanwhile, probabilistic constraint satisfaction encodes safety via distributional assumptions, but does not allow the ego player to anticipate more specific patterns of adversarial interactions with non-ego agents.

\begin{figure}
    \centering
    \includegraphics[scale=0.36]{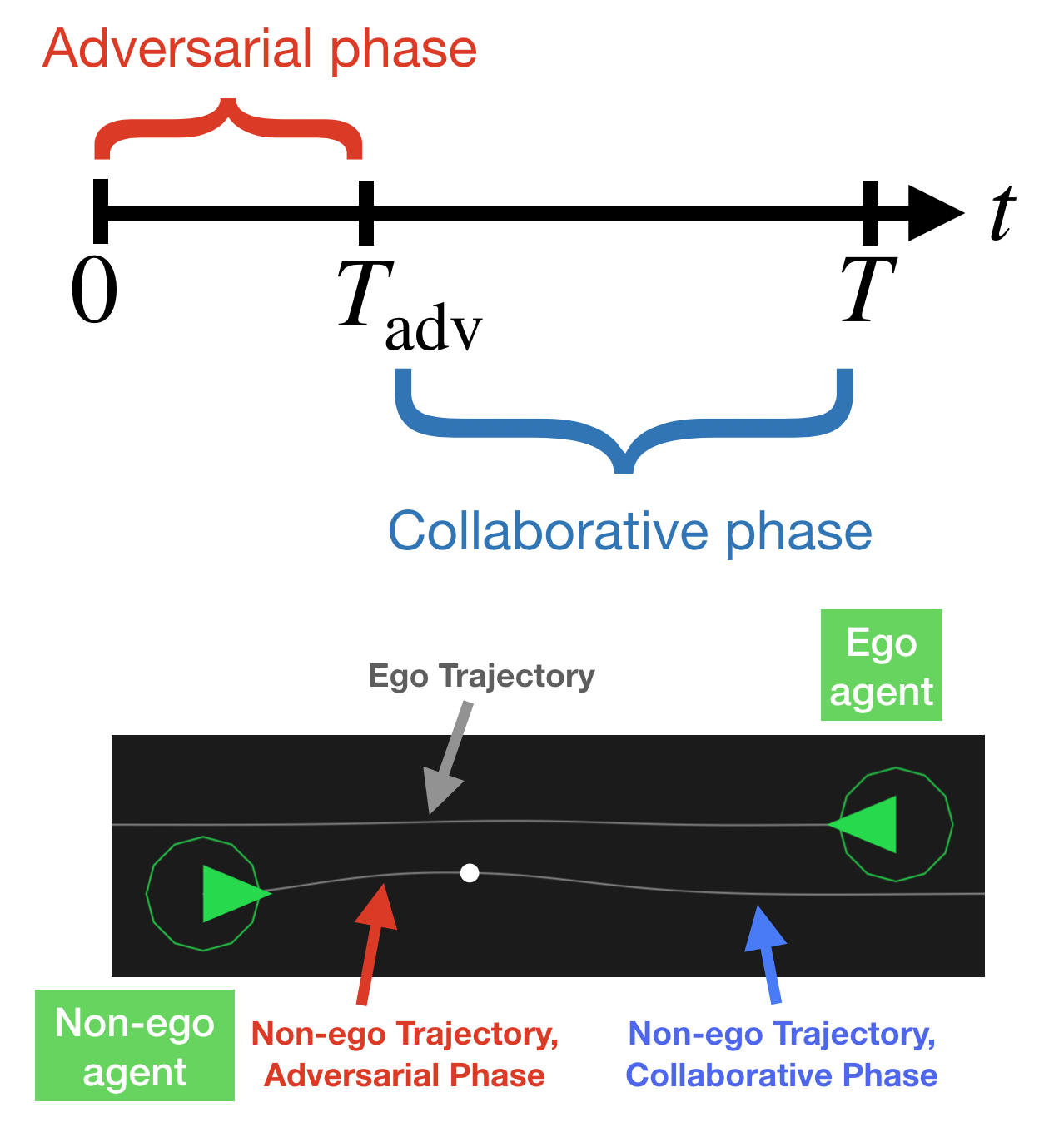}
    \caption{\emph{(Top)} To encode robustness into its safety guarantees, the ego agent imagines that all other agents behave adversarially during an initial time frame before resuming ``normal'' cooperative behavior.
    \emph{(Bottom)} In this example, an ego vehicle assumes that the non-ego, oncoming vehicle behaves adversarially for an initial period of time, which results in its swerving into the ego's lane. After the adversarial time interval expires, the non-ego agent is assumed to return to its own lane, and resume cooperative behavior for the remainder of the time horizon.  \label{fig:front}}
\end{figure}

This paper addresses these issues by presenting a third, novel formulation of robustness in multi-agent games, illustrated in \figref{fig:front}. Here, the ego agent presupposes that, for an initial fixed window of time $\advhorizon$, other agents are temporarily ``distracted'' and thus behave erratically, but revert to normal operation afterwards for the rest of the time horizon $T$. 
The ego agent encodes this behavior in a dynamic game by modifying other agents' cost structure, and solving the resulting game in a receding time horizon. 
Once computed, the ego agent begins to execute its equilibrium strategy, and after a small time has elapsed, the ego agent re-solves the game accounting for updated initial conditions.
More specifically, we encode this notion of ``temporary distraction'' in each non-ego agent’s cost function using time-varying running cost terms defined over two stages---an \textit{adversarial phase}, during which the ego agent presumes that other agents are deliberately choosing control strategies to endanger the ego's safety, followed by a \textit{cooperative phase}, during which the non-ego agents are expected to resume \say{normal} cooperative behavior. This is modeled on the concept of defensive driving, wherein a driver on a busy road guards themselves against other drivers who, while momentarily distracted, may temporarily behave dangerously.

The rest of our paper is structured as follows. \secref{sec:related_work} presents related work on the use of HJI reachability theory 
and chance-constrained optimal control to encode adversarial and probabilistic robustness, respectively, as well as recent literature on iterative algorithms for solving dynamic games. \secref{sec:preliminaries} presents the mathematical formulation for the dynamic game that models the multi-agent interactions studied in our work. \secref{sec:methods} presents the main methodology in this paper, and formally introduces the concept of an adversarial time horizon. \secref{sec:results} demonstrates the spectrum of robustness which can be expressed in our formulation, compared with a purely cooperative game-theoretic approach, using multiple traffic scenarios. \secref{sec:discussion} summarizes our contributions and discusses directions for future research.

\section{Related Work}
\label{sec:related_work}

This section reviews prior literature on several of the prevailing formulations of safety used in the design of multi-agent and uncertain systems---\textit{adversarial reachability}, \textit{multi-agent forward reachability}, and \textit{probabilistic constraint satisfaction}. We compare and contrast these against the novel methodology in our work. We conclude this section with a brief summary of modern algorithms for iteratively solving dynamic games.

\subsection{Adversarial Reachability}

Adversarial reachability methods \cite{evans1984differential, mitchell2005time, fisac2015pursuit, bansal2017hamilton} involve the construction of a zero-sum differential game between two agents, the ego agent and an adversary. The Nash equilibrium of this game satisfies a Hamilton-Jacobi-Isaacs (HJI) partial differential equation, which can be numerically solved via state space discretization. In this zero-sum framework, the ego agent assumes that the adversarial agent is constantly attempting to compromise the ego's safety, and will thus compute and execute a control strategy that steers the ego's trajectory away from any feasible trajectory of the non-ego agent. 

Adversarial reachability appropriately describes many intricate dynamic interactions, such as capture-the-flag, reach-avoid, and pursuit-evasion-defense games \cite{isaacs1951games, isaacs1999differential, fisac2015pursuit}. However, it suffers from several significant limitations. First, this zero-sum formulation can only model dynamic interactions between two agents, or two groups of colluding agents, with opposing goals. 
This is inadequate for many motion-planning tasks, such as those involving traffic scenarios, which must account for the presence of an arbitrary number of agents, with possibly an arbitrary number of goals. Second, the adversarial nature of the zero-sum game leads to the construction of extremely conservative ego trajectories, since the ego must imagine the worst-case non-ego behaviors that can possibly transpire \figref{fig:front}. Our approach, on the other hand, avoids the first issue by considering a general-sum game applicable to $N$-player scenarios. Moreover, we avoid the second issue by modeling antagonistic non-ego behavior using the novel notion of an  adversarial-to-cooperative time horizon, as shown in \figref{fig:front}, rather than as a worst-case bounded disturbance. By modeling non-ego agents as first adversarial, then cooperative, we avoid overly conservative ego strategies corresponding to purely adversarial non-ego trajectories that are unlikely to materialize.

\subsection{Multi-Agent Forward Reachability}
\label{subsec:multi-agent forward reachability}

Reachability-based formulations can also be used for safety-critical path planning in a non-game theoretic manner. For instance, forward reachable sets (FRS) of the ego agent can be computed offline by numerically solving the Hamilton-Jacobi-Bellman equation, then used to aid online motion planning modules in the generation of obstacle-avoiding trajectories.
This is the approach taken by the Reachability-based Trajectory Design for Dynamical environments (RTD-D) and RTD-Interval (RTD-I) algorithms presented in \cite{Vaskov2019NotAtFaultDrivinginTraffic, Yu2020RiskAssessmentAndPlanning}, in which not-at-fault ego trajectories are generated by leveraging the offline-computed FRS of the ego agent and online obstacle motion predictions from an external module. Although the resulting trajectories avoid at-fault collisions, this framework does not allow the ego agent to account for the dynamic reactions of non-ego agents to its behavior. In contrast, our work explicitly models the dynamic obstacles as non-ego agents, within the framework of a dynamic feedback game.

Geometric prediction modules form another framework for using reachability-based methods in a non-game-theoretic setting. For instance, \cite{Koschi2018SetBasedPO, Pek2018ComputationallyEF} ensure constraint satisfaction by computing fail-safe ego trajectories which avoid an overapproximation of all dynamically feasible non-ego trajectories. This is posed as an optimal control problem (rather than a dynamic game), with avoidance of the set of all feasible non-ego trajectories serving as a state constraint. Although these approaches ensure that the ego will not collide with the non-ego agent, they do so at the cost of generating overly conservative maneuvers, particularly in situations when the non-ego agent may not be purely adversarial throughout the entire time horizon. By contrast, our formulation generates less conservative trajectories by assuming that non-ego agents display hostile behavior only during a fixed subset of the overall time horizon.

\subsection{Probabilistic Constraint Satisfaction}

Probabilistic constraint satisfaction is another commonly used framework for establishing safety guarantees in motion planning. These approaches bound the probability that an ego agent, operating in an unpredictable environment with stochastic disturbance, becomes unsafe \cite{OnoBlackmore2010ChanceConstrainedFiniteHorizonOptimalControl}. In particular, risk-sensitive algorithms guard the ego agent from low-probability events that may result in highly dangerous outcomes. For example, \cite{farshidian2015RiskSensitiveNonlinear} generates risk-sensitive trajectories by optimizing an exponential-quadratic cost term that amplifies the cost of low-probability, yet highly dangerous outcomes. Meanwhile, \cite{pilipovsky2020ChanceConstrainedOptimal} associates individual constraint violations with different penalties, and optimizes their allocation over the horizon. Similarly, \cite{du2012robot} uses a probabilistic framework to capture state uncertainties of an autonomous robotic agent, and uncertainties in the geometry or dynamics of obstacles in its environment. However, the control strategies generated by these methods merely account for the nonzero probability of unsafe outcomes occurring any time within the entire time horizon. By contrast, our work allows the ego to explicitly encode adversarial non-ego behavior inside a specific subset of the time horizon, when such behavior is most expected to occur.

\subsection{Algorithms for Solving Dynamic Games}


Several alternative methods exist in the literature for numerically solving the general-sum differential games considered in this work. For instance, one can derive a set of coupled Hamilton-Jacobi equations whose solutions yield local Nash equilibrium strategies \cite{starr1969nonzero, starr1969further}, which can then be computed numerically via state space discretization. However, these algorithms inevitably require computational cost and memory that scale exponentially with the state dimension (the \say{curse of dimensionality}), and are thus unsuitable for modeling the high-dimensional, multi-player interactions considered in our paper \cite{bertsekas2001DynamicProgrammingandOptimalControl}.

Another category of numerical methods for dynamic games involves Iterative Best Response (IBR) algorithms \cite{fisac2018hierarchical, wang2019game}. These algorithms iterate through the players, repeatedly solving the optimal control problem of finding the best-response strategy of each player, assuming all other players' strategies are currently fixed at the values in the previous iterate. Replacing the full dynamic game with a sequence of optimal control problems reduces computation time at each iteration; however, IBR algorithms can still be computationally inefficient overall.


Our work uses ILQGames \cite{fridovich2019efficient}, a recently developed iterative linear-quadratic algorithm, as our primary game solver. ILQGames iteratively solves linear-quadratic games and incurs computational complexity cubic in the number of players and linear in the time horizon. 
Although other game solvers, such as the Augmented Lagrangian GAME-theoretic Solver (ALGAMES) \cite{cleac2019algames}, exist, all known implementations of these methods are restricted to an open-loop \cite{di2019newton, di2020local}---rather than feedback---information structure and have equivalent computational complexity. 



    

    

    

\section{Preliminaries}
\label{sec:preliminaries}


Consider the $N$-player finite horizon general-sum differential game with deterministic, noise-free nonlinear system dynamics:
\begin{equation}
    \dot{x} = f(t, x, u_{1:N}).
\end{equation}
Here, $x \in \R^n$ is the state of the system, obtained by concatenating the dynamical quantities of interest of each player, $t \in \R$ denotes time, $u_i \in \R^{m_i}$ is the control input of player $i$, for each $i \in \{1, \cdots, N\} := [N]$, and $u_{1:N} := (u_1, \cdots, u_N) \in \R^m$, where $m := \sum_{i=1}^N m_i$. The dynamics map $f: \R \times \R^n \times \R^m \ra \R^n$ is assumed to be continuous in $t$ and continuously differentiable in $x$ and $u_i$, for each $i = 1, \cdots, N$ and each $t \in [0, T]$. 
Since we wish to ensure the safety of one particular player amidst their interactions with all other players, we refer to Player 1 as the \textit{ego agent}, and the other players as \textit{non-ego agents}.
Each player's objective is defined as the integral of a running cost $\runningcost_i: [0, \horizon] \times \R^n \times \R^m \ra \R$ over the time horizon $[0, \horizon]$:
\begin{align} \label{Eqn: Running Cost}
    \cost_i\big(\control_{1:\numplayers}(\cdot) \big) = \int_0^T \runningcost_i\big(t, \state(t), \control_{1:\numplayers}(t) \big) \hspace{0.5mm} dt,
\end{align}
for each $i \in \{1, \cdots, \numplayers\}$. The running costs $\runningcost_i$ encode implicit dependence on the state trajectory $\state(\cdot): [0, T] \ra \R^n$ and explicit dependence on the control signals $\control_{i}(\cdot): [0, \horizon] \ra \R^m$.

To minimize its cost, each player selects a control strategy to employ over the time horizon $[0, \horizon]$, as described below. We assume that, at each time $t \in [0, \horizon]$, each player $i$ observes the state $x(t)$, but no other control input $\{\control_j(t) \mid j \ne i\}$, and uses this information to design their control, i.e.
\begin{align*}
    \control(t) := \feedback_i(t, \state(t)),
\end{align*}
where $\feedback_i: [0, T] \times \R^n \ra \R^{m_i}$, defined as Player $i$'s \textit{strategy}, is assumed to be measurable. We define the \textit{strategy space of Player $i$}, denoted $\feedbackset_i$, as the collection of all of Player $i$'s possible strategies, and denote, with a slight abuse of notation, the overall cost $\cost_i$ of each Player $i$ by: 
\begin{align*}
    \cost_i(\feedback_1; \cdots; \feedback_\numplayers) 
    := \cost_i\big(\feedback_1(\cdot, \state(\cdot)), \cdots, \feedback_\numplayers(\cdot, \state(\cdot)) \big).
\end{align*}
In practice, we shall solve for strategies $\feedback_i$ that are time-varying, affine functions of $\state$. 

We now define the \textit{Nash equilibrium} of the above game.

\begin{definition}
(Nash equilibrium, \cite[Chapter 6]{basar1999dynamic})
The strategy set $\left(\gamma_{1}^\star, \cdots, \gamma_{N}^\star \right)$ is said to be a Nash equilibrium if no player is unilaterally incentivized to deviate from his or her strategy. Precisely, the following inequality must hold for each player $i$:
\vspace{-0.25cm}
\begin{align} \label{Eqn: Nash Equilibrium}
\cost_{i}^\star & := \cost_{i}\left(\feedback_{1}^\star, \ldots, \feedback_{i-1}^\star, \feedback_{i}^\star, \feedback_{i+1}^\star, \ldots, \feedback_{\numplayers}^\star\right) \\ \nonumber
& \leq \cost_{i}\left(\feedback_{1}^\star, \ldots, \feedback_{i-1}^\star, \feedback_{i}, \feedback_{i+1}^\star, \ldots, \feedback_{\numplayers}^\star\right), \forall \feedback_{i} \in \feedbackset_{i}.
\end{align}
\end{definition}

Computing a global Nash equilibrium is intractable for dynamic games with general dynamics and cost functions. As such, in this work, we concern ourselves with finding a \textit{local} Nash equilibrium, which is defined similarly to \eqref{Eqn: Nash Equilibrium}, but with the inequalities only constrained to hold within a neighborhood of the strategy set $(\gamma_1^\star, \cdots, \gamma_N^\star)$. Moreover,
we impose
additional constraints 
on the dynamical quantities of each player, to model appropriate behavior between autonomous agents in traffic scenarios. These constraints will translate into a set of \textit{state constraints}, and will significantly affect the set of Nash equilibria of the game. As such, in this work, we search for a (similarly defined) \textit{generalized local Nash equilibrium}.

\section{Methods}
\label{sec:methods}

Our main contribution is a novel formulation of safety, best understood through the lens of defensive driving. In \secref{subsec:defensive driving}, we describe how, in the ego agent's mind, the concept of defensive driving can be encoded into the running cost of each non-ego agent, i.e. $g_i(x, u_{1:N})$, for each $i \in \{2, \cdots, N\}$. To demonstrate this defensive driving framework in practice, we simulate realistic traffic scenarios; \secref{subsec:simulation_setup} details the dynamics, costs, and constraints imposed on the various agents in these simulations. Finally, in \secref{subsec:ILQgames}, we summarize the ILQGames algorithm as the main feedback game solver used in this work.

\subsection{Encoding Defensive Driving as a Running Cost}
\label{subsec:defensive driving}

In our framework, the ego agent (Player 1) operates under the assumption that all other agents are momentarily distracted. To encode this \say{imagined} scenario, the ego agent imagines the overall time horizon $[0, T]$ as divided into two sub-intervals, the \textit{adversarial interval} $[0, \advhorizon]$ and \textit{cooperative interval} $[\advhorizon, T]$, with $0 < \advhorizon < T$. During the adversarial interval, the ego agent imagines the other agents to be \say{momentarily distracted,} and desires to act \textit{defensively}. This phenomenon is modeled using an adversarial running cost $\advrunningcosti: \R^\xdim \times \R^{\numplayers \udim} \ra \R$ for each $i \in \{2, \cdots, N\}$. On the other hand, during the cooperative interval, the ego agent supposes that the other agents have reverted to their \say{normal} or \say{cooperative} manner, and thus proceeds to select control signals for the remainder of the time horizon in a less conservative manner. This behavior is captured using a cooperative running cost $\cooprunningcosti: \R^\xdim \times \R^{\numplayers \udim} \ra \R$ for each $i \in \{2, \cdots, N\}$. In other words, the running cost of each non-ego agent $\runningcost_i$ can be piecewisely defined as follows: 
\begin{align*}
    \runningcost_i(t, \state, \control_{1:N}) = \begin{cases}
    \advrunningcosti(\state, \control_{1:N}), \hspace{1cm} &t \in [0, \advhorizon), \\
    \cooprunningcosti(\state, \control_{1:N}), &t \in [\advhorizon, T].
    \end{cases}
\end{align*}
In this scenario, the net integrated cost $J_i$, first defined in \eqref{Eqn: Running Cost} can be written as follows:
\begin{equation}
  \label{eqn:split_horizon_cost}
  J_i = \int_0^{\advhorizon} \advrunningcosti(\state, \control_{1:\numplayers}) dt + \int_{\advhorizon}^{\horizon} \cooprunningcosti(\state, \control_{1:\numplayers}) dt\,.
\end{equation}
With increasing $\advhorizon$, the ego agent imagines an increasingly adversarial encounter and acts more and more defensively as a result. In practice, the user or system designer would select a suitable $\advhorizon$ before operation, e.g., by choosing the largest $\advhorizon$ such that the solution deviates from a nominal solution with $\advhorizon = 0$ sufficiently little.

\subsection{Simulation Setup}
\label{subsec:simulation_setup}

To test this construction, we simulate two traffic encounters in ILQGames \cite{fridovich2019efficient} that involve significant interaction (see Sec.~\ref{sec:results}), in which a responsible human driver would likely drive defensibly. Our method attempts to capture the spectrum of this ``defensive'' behavior as $\advhorizon$, the duration of the adversarial time horizon, is varied. In each setting, each agent (in this case, each car) has augmented bicycle dynamics, i.e.:
\begin{align}
    \label{eqn:bicycle_dyn}
    \dot p_{x, i} &= v_i \sin \theta_i, \hspace{1cm}&\dot v_i = a_i, \nonumber\\
    \dot p_{y, i} &= v_i \cos \theta_i, \hspace{1cm}&\dot \phi_i = \omega_i, \\
    \dot \theta_i &= (v_i / L_i) \tan \phi_i, \hspace{1cm}&\dot a_i = j_i, \nonumber
\end{align}
where $\state = (p_{x, i}, p_{y, i}, \theta_i, v_i, \phi_i, a_i)_{i=1}^\numplayers$ encapsulates the position, heading, speed, front wheel angle, and acceleration of all vehicles, ${\control_i = (\omega_i, j_i)}$ represents each vehicle's front wheel rate and tangent jerk, and $L_i$ is each player's inter-axle distance.

We define $\advrunningcosti$ and $\cooprunningcosti$ as weighted combinations of the following functions, with different behavior encouraged through the use of different weighting coefficients. 
We denote $p_i = (p_{x, i}, p_{y, i})$ for each agent position, $d_{\lane_i}(p_i)$, defined below, for the distance between an agent and the corresponding lane centerline $\lane_i$ in the $(p_{x, i}, p_{y, i})$-plane, and $\thresh_{\textnormal{prox}}$ for a constant desired minimum proximity between agents:
\begin{align}
    \text{lane center:}~&\left[d_{\lane_i}(p_i) := \min_{p_\lane \in \lane_i} \|p_\lane - p_i\|\right]^2 \label{eqn:lane_center} \\
    \text{ideal speed:}~&(v_i - v_{\textnormal{ref}, i})^2 \label{eqn:ideal_speed}\\
    \text{cooperative:}~&\mathbf{1}\{\|p_i - p_j\| < \thresh_{\text{prox}}\} (\thresh_{\textnormal{prox}} - \|p_i - p_j\|)^2  \label{eqn:coop_cost}\\
    \text{adversarial:}~&\|p_i - p_j\|^2  \label{eqn:adv_cost}\\
    \text{input:}~&u_i^T R_{ii} u_i \label{eqn:input_cost}\,.
\end{align}

\begin{figure*}[tbp]
\centering
\includegraphics[scale=0.6, width=0.8\textwidth]{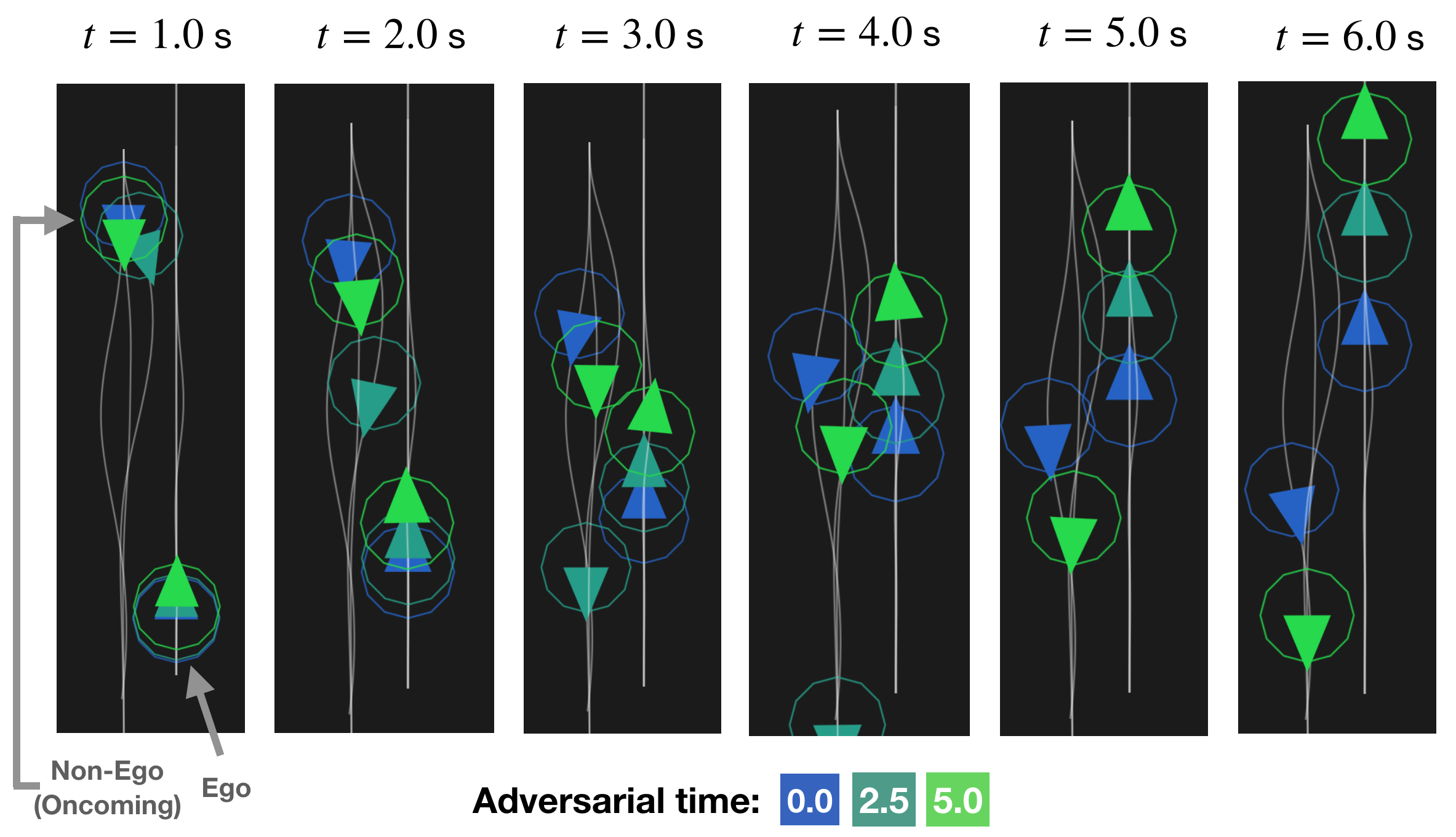}
\caption{\textbf{Oncoming example.} The ego vehicle (right lane, heading upwards) and the oncoming vehicle (left lane, heading downwards) perform increasingly extreme maneuvers as  $\advhorizon$ increases, in this ``oncoming'' scenario. Dark blue, turquoise, and light green are used to represent the agents' location for $\advhorizon = 0, 2.5, \SI{5}{\second}$, respectively. Panels show agent positions as time elapses. When $\advhorizon = \SI{0}{\second}$, the ego vehicle does not deviate significantly from its lane from $t = 1$ to $\SI{6}{\second}$ because it anticipates that the non-ego vehicle will behave cooperatively throughout the entire time horizon by swerving to avoid a collision.  However, when $\advhorizon = \SI{5}{\second}$, the ego vehicle actively swerves outward to avoid the non-ego agent. This minor course adjustment is sufficient to dissuade the oncoming vehicle from making a stronger effort to attempt a collision, since it would occur in the future, after $\advhorizon$.}
\label{fig:oncoming_panel}
\end{figure*}

\afterpage{
\begin{figure*}[tbp]
\centering
\includegraphics[scale=0.4]{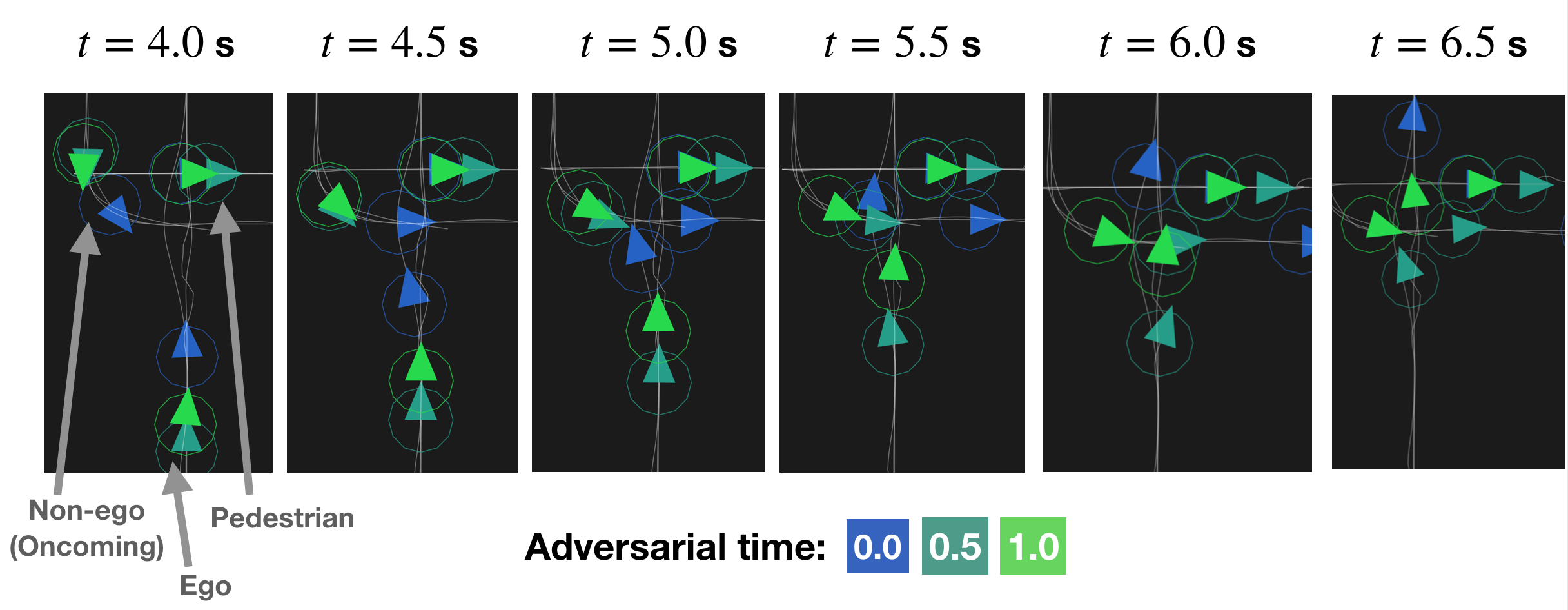}
\caption{\textbf{Three Player Intersection example.}
The ego vehicle (right lane, heading upwards) navigates an intersection while avoiding collision with an oncoming vehicle (left lane, heading downwards initially before making a left turn) and a pedestrian (horizontal path at the intersection, left to right). Dark blue, turquoise, and light green are used to represent the agents' location at $\advhorizon = 0, 0.5, 1$ s, respectively. As with the oncoming example, these three agents perform increasingly extreme maneuvers as $\advhorizon$ increases. In particular, when $\advhorizon = 0$ s, the ego vehicle anticipates that the non-ego will continue along its curved path at its nominal speed, allowing it to approach the intersection before the ego vehicle. Thus, the ego vehicle swerves leftwards, to avoid the non-ego agent as it makes its left turn and continues rightwards in the figure, resulting in a collision-free trajectory. (The pedestrian's trajectories for $\advhorizon = 0$ s and $\advhorizon = 0.5$ s coincide.) However, when $\advhorizon = 1$ s, the ego vehicle accelerates and swerves rightwards to avoid the non-ego vehicle. As before, this suffices to dissuade the oncoming vehicle from performing more collision-seeking behavior, since it forces any potential collision to occur after $\advhorizon$. The pedestrian's speed also noticeably decreases for the same reason.}
\label{fig:three_player_intersection_panel}
\end{figure*}}

Recall that, for non-ego agents, the ``adversarial'' cost is only present during the adversarial horizon $[0, \advhorizon)$ and the ``cooperative'' cost is present thereafter during the cooperative horizon $[\advhorizon, T]$. We also enforce the following inequality constraints, where $\thresh_{\textnormal{lane}}$ denotes the lane half-width, and $\underline{v}_i$ and $\overline{v}_i$ denote speed limits:
\begin{align}
    \text{proximity:}~&\|p_i - p_j\| > \thresh_{\textnormal{prox}} \label{eqn:prox_constraint}\\
    \text{lane:}~&|d_{\lane_i}(p_i)| < \thresh_{\textnormal{lane}} \label{eqn:lane_constraint}\\
    \text{speed range:}~&\underline{v}_i < v_i < \overline{v}_i \label{eqn:speed_constraint}\,,
\end{align}
Here, the ``proximity'' constraint is enforced for only the ego agent, to force the ego to bear responsibility for satisfying joint state constraints which encode his or her own safety (e.g. non-collision). In addition, all agents must satisfy individual constraints that encode reasonable conduct in traffic (e.g., staying within a range of speeds). All constraints are enforced over the entire time horizon $[0, T]$. For all tests, we use a time horizon $\horizon = \SI{15}{\second}$ and discretize time (following \cite{fridovich2019efficient} and \cite{basar1999dynamic}) at $\SI{0.1}{\second}$ intervals. 




\subsection{Implementation Details}
\label{subsec:ILQgames}

The traffic simulations in this work are solved approximately to local feedback Nash equilibria in real time using ILQGames, a recently developed, open-source C++ based game solver algorithm introduced in \cite{fridovich2019efficient}. ILQGames iteratively solves linear-quadratic games, obtained by linearizing dynamics and quadraticizing costs, and incurs computational complexity that is cubic in the number of players \cite{fridovich2019efficient}. 
As discussed above, we must also account for both equality and inequality constraints on the game trajectory. 
While we note that \cite{fridovich2019efficient} does not address constrained Nash games, here we incorporate constraints using augmented Lagrangian methods \cite{nocedal2006numerical}. For a more detailed discussion of constraint-handling in feedback Nash games, please refer to \cite{laine2021computation}. 
Note that although other game solvers, such as ALGAMES \cite{cleac2019algames} and Iterative Best Response algorithms \cite{wang2019game}, can likewise handle constraints, their applications are restricted to open-loop games. A thorough treatment of constraints in games can be found in \cite{Peng2011NonconvexGameswithSideConstraints}. 


\section{Results}
\label{sec:results}



We present simulation results for various traffic scenarios in which a responsible traffic participant would likely drive defensively. First, we consider a simple situation involving oncoming vehicles on a straight road, as a proof of concept. Then, we analyze a more complicated intersection example with a crosswalk. In both cases, the ILQGames algorithm solves the defensive driving game quickly, in under $\SI{1}{\second}$.

\subsection{Oncoming Example}
\label{subsec:oncoming}

In this example, the ego car is traveling North on a straight road when it encounters another car traveling South. Since the road has a lane in each direction, ``ideally'' the ego vehicle would not deviate too far from its lane or speed. However, to drive more defensively, the ego vehicle should plan as though the oncoming Southbound car were to act noncooperatively. Our method encodes precisely this type of defensive planning. \figref{fig:oncoming_panel} shows the planned trajectories that emerge for increasing $\advhorizon$. As shown, the ego vehicle (bottom) imagines more aggressive maneuvers for itself and the oncoming car (top) as $\advhorizon$ increases. Note, however, that these are merely \emph{imagined} trajectories and that (a) the ego vehicle can always choose to follow this trajectory only for an initial period of time, and recompute its trajectory thereafter,
 and (b) the oncoming vehicle will make its own decisions and will \emph{not} generally follow this ``partially adversarial'' trajectory. We solve each of these problems (with fixed $\advhorizon$) in under $\SI{0.5}{\second}$.

\subsection{Three-Player Intersection Example}
\label{subsec:3-player intersection, receding horizon}


We introduce a more complicated scenario designed to model the behavior of two vehicles and a pedestrian at an intersection. As shown in \figref{fig:three_player_intersection_panel},
the ego vehicle is present in the intersection alongside a non-ego vehicle heading in the opposite direction, who wishes to make a left turn, and a pedestrian, who wishes to cross the road. 
To reach their goal locations, these three agents must cross paths in the intersection.
When $\advhorizon = 0$ s, the ego vehicle continues straight along its lane because it anticipates that the non-ego vehicle will behave cooperatively throughout the entire time horizon. 
In particular, it anticipates that the non-ego vehicle will continue along its curved path at nominal speed, resulting in a collision-free trajectory. 
However, as with the oncoming example, the ego vehicle's trajectory becomes increasingly more conservative as the adversarial time horizon increases in length. 
When $\advhorizon = 1$ s, the ego vehicle actively swerves rightwards to avoid the non-ego vehicle. 
This is because in this scenario, the oncoming vehicle is initially slower than the ego vehicle, and will thus approach the intersection at the same time as the ego vehicle. 
As before, each problem is solved in well under 
$\SI{0.75}{\second}$ in single-threaded operation on a standard laptop,
via the ILQGames algorithm \cite{fridovich2019efficient}.
This performance indicates real-time capabilities which will be explored in future work on hardware. 

\section{Discussion}
\label{sec:discussion}

We have presented a novel formulation of robustness in motion planning for multi-agent problems. Inspired by defensive driving, our method explicitly models other agents as adversarial in only a limited, initial portion of the overall planning interval. Like adversarial methods in Hamilton-Jacobi-Isaacs (HJI) optimal control, our approach draws upon earlier work in differential game theory. However, instead of forcing the ego to 
avoid 
all feasible non-ego trajectories, we use a piecewise-defined game cost structure to endow the ego with the perspective that other agents are temporarily distracted. As such, our approach generates far less conservative behavior  than purely adversarial 
methods. Simulation results illustrate that this novel formulation of safety can be used to solve these ``defensive'' problems in real-time. We are eager to implement this method in hardware and test its performance in a receding time horizon with other (human) drivers.

\section{Future Work}
\label{sec:future_work}

Future work will examine more flexible formulations for encoding defensive behavior in games. 
For example, in many real-life scenarios, vehicle dynamics should be modeled with stochastic, partial observations, and should account for occlusions and dynamic disturbances. 
Also, the ego agent may wish to select the adversarial time horizon more flexibly. 

Of particular interest are cases where the ego agent may choose to vary $\advhorizon$ from one non-ego agent to another.
For example, the ego may observe that some non-ego agents are behaving more adversarially than others, and respond accordingly by associating such players with higher values of $\advhorizon$, compared to other non-ego agents. In addition, the ego may wish to allocate different parts of the overall time horizon to be adversarial, rather than simply the first $\advhorizon$ seconds. For example, choosing the adversarial time horizon to be the final $\advhorizon$ seconds of the overall time horizon, rather than the first $\advhorizon$ seconds, would transform the game from an adversarial-to-cooperative type to a cooperative-to-adversarial type. This would be useful in situations where the ego agent predicts that the surrounding non-ego agents are currently cooperative, but may become momentarily distracted in the near future. 
For example, such a formulation may cause the ego agent to gradually approach an intersection at which other agents might run a red light. 


Future work will also examine the physical implementation of this framework on hardware, and test its performance in a receding horizon fashion in real-life traffic scenarios with other human drivers. We also anticipate using this novel notion of safety in game-theoretic applications outside of motion planning, such as in economics or public health applications, where it may be prudent to model certain agents as adversarial in certain time intervals.







\balance
\printbibliography

\end{document}